\documentclass[12pt]{article}

\usepackage{amssymb,amsmath,amsfonts,latexsym,graphicx}
\usepackage[numbers]{natbib}
\usepackage{multirow}
\usepackage{hyperref}
\usepackage{xcolor}
\usepackage{float}

\hypersetup{
    colorlinks=true,
    urlcolor=blue,
    citecolor=black,
    pdftitle={EndoInsight After Competition Results},
    pdfpagemode=FullScreen,
}

\begin{document}

\begin{center}

\Large \bf ResNet-50 with Structured Localization-First Temporal Decoding for Gastrointestinal Video Analysis \textcolor{blue}{- after competition results} \rm

\vspace{1cm}

\large Romil Imtiaz$\,^a$, \large Dimitris K. Iakovidis$\,^a$

\vspace{0.5cm}

\normalsize

$^a$ Department of Computer Science and Biomedical Informatics, University of Thessaly, Papasiopoulou 2--4, Lamia 35131, Greece

\vspace{5mm}

Corresponding Author Email: {\tt \href{mailto:rimtiaz@uth.gr}{rimtiaz@uth.gr}}

Team Name: {\tt EndoInsight}

Updated GitHub Repository Link: {\tt \href{https://github.com/innoisys/ResNet-50-for-GI-Events}{https://github.com/innoisys/ResNet-50-for-GI-Events}}

\vspace{1cm}

\end{center}

\begin{abstract}
We developed a gastrointestinal video analysis pipeline based on a ResNet-50 frame-level classifier followed by anatomy-guided temporal event decoding. The frame-level model predicts anatomical and pathological labels from gastrointestinal video frames, while the temporal stage converts framewise predictions into event-level outputs aligned with the official evaluation format. During the competition, our system used clipped class-wise positive weighting to address severe class imbalance and anatomy-guided temporal refinement to improve event consistency. After the competition, we updated the temporal decoding pipeline by replacing flat per-label decoding with a structured localization-first sequence decoder. The updated decoder first uses the predicted anatomical region as the main temporal guide, applies vote-based smoothing and progression locking to enforce realistic gastrointestinal tract order, and then decodes sub-localization and pathology labels under anatomy-based compatibility constraints. This post-competition update achieved an average mean Average Precision at 0.5 temporal IoU threshold (mAP@0.5) of 0.4313 and mean Average Precision at 0.95 temporal IoU threshold (mAP@0.95) of 0.4118 across the three evaluated videos, improving the stricter mAP@0.95 score from 0.4020 to 0.4118.
\end{abstract}

\section{Motivation}\label{sec1}

Automatic analysis of gastrointestinal videos is important for computer-assisted diagnosis, lesion screening, and efficient clinical review of long endoscopic examinations. In gastrointestinal video analysis, the main challenge is not only to recognize abnormal visual patterns, but also to localize them accurately over time. Rare findings may appear briefly, coexist with changing anatomical regions, and be under-represented in the training data. Therefore, the task requires both robust frame-level recognition and temporally consistent event construction.

Our motivation for participating in the competition was to investigate how a stable visual backbone, combined with medically informed temporal decoding, can improve event-level gastrointestinal video understanding. In particular, we focused on addressing two major issues: severe class imbalance among rare pathology labels and the mismatch between frame-level predictions and the official event-level ground-truth structure. The challenge provided a useful setting to study how anatomical priors, temporal smoothing, and structured event composition can improve clinically meaningful video-level predictions.

\section{Methods}\label{sec2}

Our pipeline consisted of a frame-level classifier followed by temporal event construction. Among the tested backbones, ResNet-50 \cite{he2016deep} provided the best balance between stability, speed, and predictive performance, and was therefore selected as the core model. Each frame was resized to $336 \times 336$ and processed independently to predict the presence of multiple gastrointestinal labels, including anatomy and pathology classes.

Let $x_i$ denote the $i$-th frame and $y_{ic} \in \{0,1\}$ the ground-truth label for class $c$. For each class, the network outputs a real-valued score $z_{ic}$ before probability normalization. This score is transformed into a probability using the sigmoid function:
\begin{equation}
p_{ic} = \sigma(z_{ic}) = \frac{1}{1+e^{-z_{ic}}},
\end{equation}
where $p_{ic}$ represents the predicted probability that class $c$ is present in frame $i$.

At the frame level, the model was trained using weighted binary cross-entropy \cite{goodfellow2016deep}. For class $c$, the loss was defined as:
\begin{equation}
L_c = -\frac{1}{N}\sum_{i=1}^{N}
\left[
w_c y_{ic}\log(p_{ic}) + (1-y_{ic})\log(1-p_{ic})
\right],
\end{equation}
where $N$ is the number of training frames and $w_c$ is the positive class weight for class $c$. The total loss over all classes was:
\begin{equation}
L = \sum_{c=1}^{C} L_c.
\end{equation}

At the temporal level, frame probabilities were converted into event predictions. A baseline hysteresis-style decoder was first used to map framewise scores into temporal segments. However, temporal debugging on the training set revealed a mismatch between predictions and the official event-level ground truth: the ground truth was fragmented whenever the full active label set changed, whereas our initial predictions were fewer and longer. To analyze this systematically, we built a dedicated debugging pipeline that generated temporal JSON files for both ground truth and predictions and reported per-video and per-label segment counts.

To reduce this mismatch, we composed final events from framewise active labels in the same style as the official ground truth. Specifically, if the active label set at frame $t$ is denoted by $S_t$, then a new temporal event is started whenever $S_t \neq S_{t-1}$.

We further stabilized anatomy predictions using a local voting window. For a frame $t$, we considered a neighborhood:
\begin{equation}
W_t = \{t-r, \ldots, t, \ldots, t+r\},
\end{equation}
where $r$ is the window radius. Let $\hat{a}_{\tau}$ be the predicted anatomy class at frame $\tau$. The final anatomy label at frame $t$ was obtained by majority voting:
\begin{equation}
a_t = \arg\max_k \sum_{\tau \in W_t} \mathbf{1}[\hat{a}_{\tau}=k].
\end{equation}

We also introduced an anatomy-guided gating mechanism for pathology predictions. The idea is that some abnormalities are anatomically plausible only in specific regions. Let $p_t^{(m)}$ denote the pathology probability for class $m$ at frame $t$, and let $A_m$ denote the set of anatomies where pathology $m$ is allowed according to our anatomy-gating prior. The gated pathology probability was defined as:
\begin{equation}
\tilde{p}_t^{(m)} =
\begin{cases}
p_t^{(m)}, & \text{if } a_t \in A_m, \\
0, & \text{otherwise}.
\end{cases}
\end{equation}

\begin{figure}[H]
    \centering
    \includegraphics[width=0.65\linewidth]{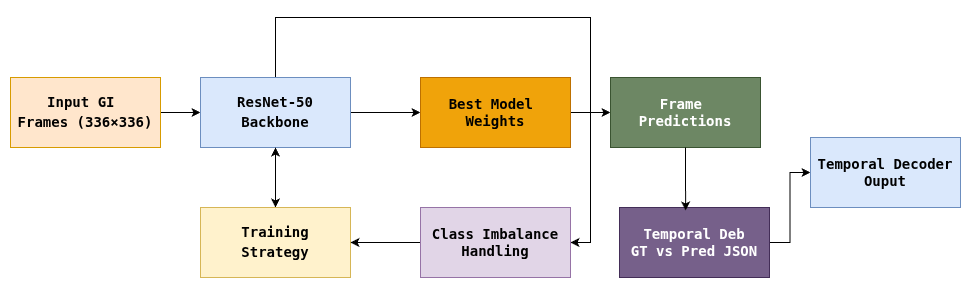}
    \caption{Overall flowchart of the proposed ResNet-50-based temporal event detection pipeline.}
    \label{fig:pipeline_after_competition}
\end{figure}

This step suppressed anatomically implausible pathology predictions and reduced false positives in regions where the target abnormality should not occur. After anatomy vote smoothing and pathology gating, the final pathology sequence was decoded using the default hysteresis-based temporal decoder. Compared with more aggressive support-window decoding, the conservative decoder generalized better on unseen data and gave the best final competition result. We also tested an HMM/Viterbi-based decoder \cite{rabiner1989hmm} with temperature scaling \cite{guo2017calibration}, but the final chosen system remained the simpler and more stable anatomy-guided hysteresis pipeline.
The overall workflow of the proposed ResNet-50-based temporal event detection pipeline is shown in Figure~\ref{fig:pipeline_after_competition}.
\subsection{How was class imbalance handled?}

Class imbalance was handled using clipped class-wise positive weighting in the binary cross-entropy loss. The positive weight for each class was computed from the ratio of negative to positive samples, increasing the learning signal for rare pathology classes. Extremely large weights were clipped to a safe range to improve rare-class learning while avoiding unstable optimization.
\subsection{\textcolor{blue}{Post-competition Changes in the Methods}}

After the competition, we updated only the temporal decoding stage while keeping the same ResNet-50 frame-level classifier. The previous temporal decoder was replaced with a structured localization-first sequence decoder. In the original frame-level output, ResNet-50 gives a probability score for each anatomy label and pathology label. For each frame, the updated decoder first selects the anatomical region with the highest anatomy probability. This selected anatomy label is then used as the main temporal guide for the rest of the decoding process.

The predicted anatomical region sequence is then refined using vote-based smoothing and progression locking. Vote-based smoothing checks each frame together with its neighboring frames and replaces short isolated anatomy changes with the dominant anatomy label in the local window. For example, if most neighboring frames are predicted as stomach but one frame is predicted as esophagus, this isolated prediction is corrected to stomach.

Progression locking then constrains the anatomy sequence to follow the realistic order of the gastrointestinal tract, such as mouth, esophagus, stomach, small intestine, and colon. If the prediction briefly moves backward or suddenly jumps to an anatomically inconsistent region, the decoder corrects this transition using the surrounding anatomical context. This helps produce a smoother and more medically plausible anatomy sequence.

After the anatomical region sequence is stabilized, sub-localization and pathology labels are decoded under anatomy-based compatibility constraints. These constraints retain a sub-localization or pathology label only when it is compatible with the predicted anatomical region. Finally, the active labels are composed framewise in the order of anatomical region, sub-localization, and pathology to generate medically consistent event predictions aligned with the official evaluation format.

\section{Results}\label{sec3}

During the competition phase, the proposed ResNet-50-based pipeline with anatomy-guided temporal decoding achieved an overall mAP@0.5 of 0.4303 and an overall mAP@0.95 of 0.4020. After the competition, the updated structured localization-first sequence decoder improved the temporal consistency of the predictions, particularly at the stricter mAP@0.95 threshold.

The post-competition results across the three evaluated videos were:
\begin{itemize}
    \item ukdd\_navi\_00051: mAP@0.5 = 0.5341, mAP@0.95 = 0.5294.
    \item ukdd\_navi\_00068: mAP@0.5 = 0.3529, mAP@0.95 = 0.3529.
    \item ukdd\_navi\_00076: mAP@0.5 = 0.4069, mAP@0.95 = 0.3529.
\end{itemize}

The average performance across the three evaluated videos was mAP@0.5 = 0.4313 and mAP@0.95 = 0.4118. The best video-level performance was obtained on ukdd\_navi\_00051, where the updated decoder achieved mAP@0.5 = 0.5341 and mAP@0.95 = 0.5294. The results show that the localization-first decoder mainly improved high-IoU temporal consistency, as reflected by the gain in mAP@0.95.

\begin{table}[H]
    \centering
    \caption{\textcolor{blue}{Performance comparison before and after the competition.}}
    \begin{tabular}{c|c|cc}
        \hline
        Video & Phase & mAP@0.5 & mAP@0.95 \\
        \hline

        \multirow{2}{*}{ukdd\_navi\_00051}
        & Before & 0.5319 & 0.5000 \\
        & After  & 0.5341 & 0.5294 \\
        \hline

        \multirow{2}{*}{ukdd\_navi\_00068}
        & Before & 0.3533 & 0.3529 \\
        & After  & 0.3529 & 0.3529 \\
        \hline

        \multirow{2}{*}{ukdd\_navi\_00076}
        & Before & 0.4056 & 0.3529 \\
        & After  & 0.4069 & 0.3529 \\
        \hline

        \multirow{2}{*}{Average (3 Videos)}
        & Before & 0.4303 & 0.4020 \\
        & After  & 0.4313 & 0.4118 \\
        \hline

        \multirow{2}{*}{Overall}
        & Before & 0.4303 & 0.4020 \\
        & After  & 0.4313 & 0.4118 \\
        \hline
    \end{tabular}
    \label{tab:competition_results}
\end{table}

\section{Discussion}\label{sec4}

The results show that improving frame-level classification alone is not sufficient for strong event-level gastrointestinal video analysis. Temporal mAP is highly sensitive to the structure of the predicted events, the ordering of labels, and the alignment between predicted segments and the official ground-truth format. Our earlier experiments showed that flat per-label decoding may produce temporally inconsistent predictions, especially when anatomical transitions and pathology labels are decoded independently.

The post-competition update addressed this limitation by introducing a structured localization-first sequence decoder. By predicting coarse anatomy first, stabilizing the anatomical trajectory, and then decoding sub-location and pathology labels under compatibility constraints, the final event predictions became more medically consistent. The improvement was most visible at mAP@0.95, suggesting that the updated decoder improved stricter temporal alignment and reduced fragmented or anatomically implausible events.

Further improvements could be achieved by integrating sequence-aware models, such as CNN-LSTM, Transformer-based temporal encoders, or end-to-end video-level training. In addition, stronger anatomical priors, uncertainty-aware event filtering, and improved rare-pathology modeling may further improve temporal event localization in future versions of the pipeline.

\section{Summary}\label{sec5}

Team EndoInsight utilized ResNet-50 as the main AI model. Clipped class-wise positive weighting was used to handle class imbalance in the dataset. Anatomy-guided temporal refinement, including anatomy vote smoothing, anatomy-based pathology gating, and GT-style framewise event composition, improved the consistency of event prediction and reduced anatomically implausible detections. The team achieved an overall mAP@0.5 of 0.4303 and an overall mAP@0.95 of 0.4020 during the competition phase.

After the competition, we improved the temporal decoding stage of the ResNet-50-based pipeline. The improvements included a structured localization-first sequence decoder, vote-based anatomical smoothing, progression locking to enforce realistic gastrointestinal tract order, and anatomy-compatible decoding of sub-localization and pathology labels. This update improved the average temporal performance across the three evaluated videos, especially at the stricter mAP@0.95 threshold. The team achieved an improved overall mAP@0.5 of 0.4313 and an overall mAP@0.95 of 0.4118.

\section{Acknowledgments}\label{sec6}

This project has received funding from the European Union's HORIZON TMA MSCA Doctoral Networks programme (HORIZON-MSCA-2023-DN-01) under Grant Agreement No. 101169012 (INTELLI-INGEST Project).

As participants in the ICPR 2026 RARE-VISION Competition, we fully complied with the competition's rules as outlined in \cite{Lawniczak2025, rarevision2026github}. Our AI model development was based exclusively on the datasets in the competition \cite{LeFloch2025figshareplus}. The mAP values were reported using the test dataset \cite{rarevision_testdata_2026}, and sanity checker \cite{manasapp} released in the competition.

\bibliographystyle{unsrtnat}
\bibliography{sample}

\end{document}